\documentclass[11pt,a4paper]{article}
\usepackage[hyperref]{emnlp2020}
\usepackage{times}
\usepackage{latexsym}

\usepackage{microtype}

\usepackage{graphicx}
\usepackage{makecell}
\usepackage{booktabs}
\usepackage{tablefootnote}
\usepackage[ruled,vlined,boxed,algo2e]{algorithm2e}
\usepackage{dblfloatfix}
\usepackage{algorithm}
\usepackage{algorithmic}
\usepackage{textcomp}
\usepackage{multirow}

\usepackage{float}

\aclfinalcopy 

\title{On the Role of Conceptualization in \\ Commonsense Knowledge Graph Construction}

\author{Mutian He$^1$, Yangqiu Song$^1$, Kun Xu$^2$, Dong Yu $^2$\\
  $^1$Hong Kong University of Science and Technology \\
  $^2$Tencent \\
  {\tt \{mhear, yqsong\}@cse.ust.hk} \\ {\tt \{kxkunxu,dyu\}@tencent.com}}

\date{}

\begin{document}
\maketitle
\begin{abstract}
  Commonsense knowledge graphs (CKGs) like \textsc{Atomic} and ASER are substantially different from conventional KGs as they consist of much larger number of nodes formed by loosely-structured text, which, though, enables them to handle highly diverse queries in natural language related to commonsense, leads to unique challenges for automatic KG construction methods. Besides identifying relations absent from the KG between nodes, such methods are also expected to explore absent nodes represented by text, in which different real-world things, or entities, may appear. To deal with the innumerable entities involved with commonsense in the real world, we introduce to CKG construction methods \emph{conceptualization}, i.e., to view entities mentioned in text as instances of specific concepts or vice versa. We build synthetic triples by conceptualization, and further formulate the task as triple classification, handled by a discriminatory model with knowledge transferred from pretrained language models and fine-tuned by negative sampling. Experiments demonstrate that our methods can effectively identify plausible triples and expand the KG by triples of both new nodes and edges of high diversity and novelty.
\end{abstract}

\section{Introduction}

Commonsense knowledge, such as knowing that {\em a trophy could not fit in a suitcase because the trophy is too big}, is implicitly acknowledged among human beings through real-life experience rather than systematic learning. As a result, artificial intelligence meets difficulties in capturing such commonsense. To deal with the issue, commonsense knowledge graphs (CKGs) like ConceptNet \citep{speer2017conceptnet}, \textsc{Atomic} \citep{sap2019Atomic}, ASER \citep{zhang2019aser}, etc. have been proposed. Such graphs are aimed at collecting and solidifying the implicit commonsense in the form of triples $\langle h,r,t \rangle$, with the head node \emph{h} and the tail node \emph{t} connected by a relation (i.e., edge) \emph{r}. However, a key difference between traditional knowledge graphs (like WordNet, Freebase, etc.) and CKGs is that commonsense is often difficult to represent as two strictly formed nodes compared by a specific relation. Instead, recent approaches represent a node with loosely-structured text, either annotated by humans or retrieved from text corpora. Nodes are then linked by one of some predefined types of relations, such as the triple $\langle$\emph{h}: I'm hungry, \emph{r}: Result, \emph{t}: I have lunch$\rangle$ in ASER.\footnote{Words in ASER are lemmatized, but in this paper we always show the original text for easier understanding.}

\begin{table}[h]
\centering
\begin{tabular}{lcc}
\toprule
& \textbf{ASER} & \textbf{\textsc{Atomic}} \\
\midrule
\textbf{\#Nodes} & 194.0M & 309.5K \\
\textbf{\#Triples} & 64.4M & 877.1K \\
\textbf{\#Relation Types} & 15 & 9 \\
\textbf{Average Degree} & 0.66 & 5.67 \\
\textbf{Entity Coverage} & 52.33\% & 6.98\% \\
\textbf{Average Distinct Entity} & 0.026 & 0.082 \\
\bottomrule
\end{tabular}
\caption{\label{Tab:CKG_size} Statistics for recently proposed commonsense knowledge graphs. Entity Coverage is calculated by the proportion of top 1\% most frequent entities in Probase, that are mentioned by nodes in each CKG. Average Distinct Entity is given by average number of distinct Probase entities per node in each CKG. Core version of ASER is used for these two results.}
\end{table}

Such CKGs, storing an exceptionally large number of triples, as shown in Table~\ref{Tab:CKG_size}, are capable of representing a much broader range of knowledge and handling flexible queries related to commonsense. However, the complexity of real world commonsense is still immense. Particularly, with innumerable eventualities involved with commonsense in the real world, it is of high cost for a CKG to cover all of them in nodes; and even when covered, to acquire the corresponding relation between each pair of nodes is of quadratic difficulty. Such a situation is particularly demonstrated by the sparsity of edges in current CKGs in Table~\ref{Tab:CKG_size}: even automatic extraction methods as in ASER, fail to capture edges between most nodes. Therefore, alternative KG construction methods are in need.


\begin{figure}[t]
\centering
\includegraphics[width=0.9\linewidth]{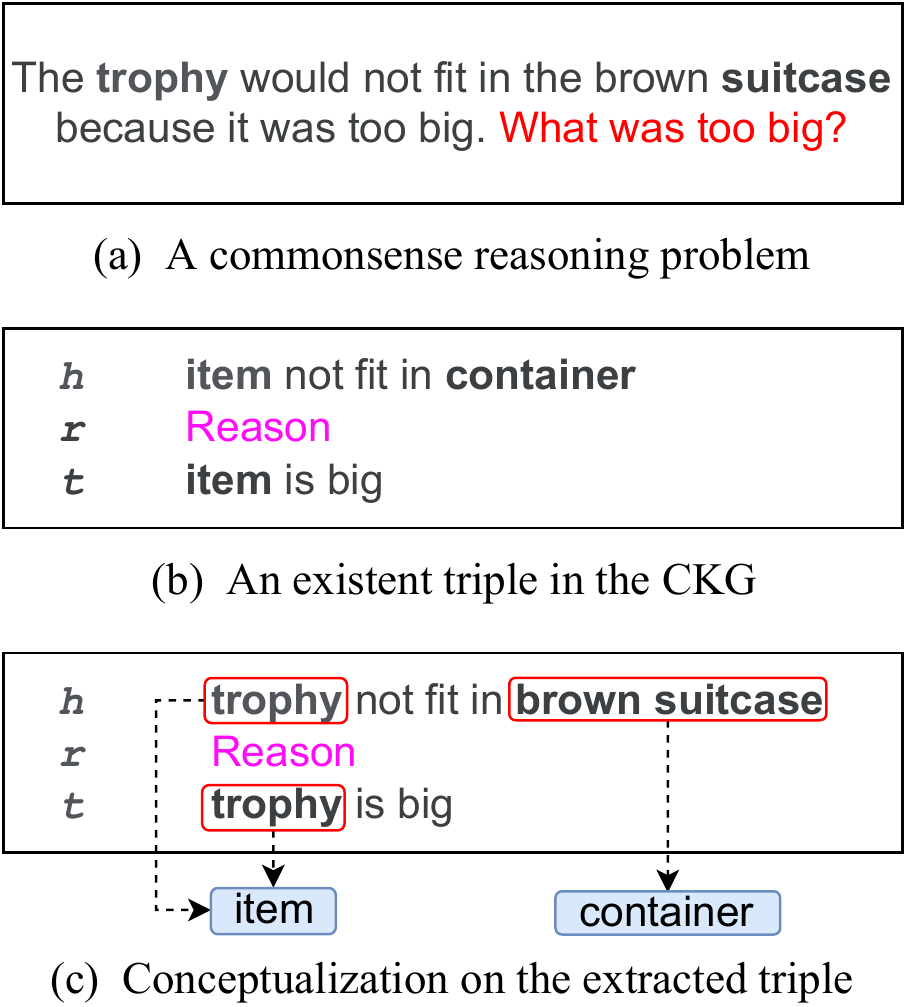}
\caption{A sample for conceptualization in CKG. Given a commonsense reasoning problem like (a), even though the corresponding triple (c) is not in the KG, a triple (b) which is present in the KG could be used through abstraction.
This is done by identifying real world entities of \textit{trophy} and \textit{brown suitcase} in the text of (c), and then substituting them using \textit{IsA} relations. Following the same idea, (c) can be produced from (b) to be included in the CKG via instantiation.}
\label{Fig:winograd}
\end{figure}

As nodes are represented by text, utilizing semantic information becomes critical for CKG construction. Such semantic information can be leveraged by large pretrained language models like BERT \citep{devlin2018bert} and GPT-2 \citep{radford2019gpt2}: 
These models capture rich semantic knowledge from unsupervised pretraining, which can be transferred to a wide range of downstream tasks to reach impressive results. 
Therefore, efforts have been made to extract knowledge from such models for KG construction. 
COMeT \citep{bosselut-etal-2019-comet} is a notable attempt which fine-tuned GPT on CKG triples to predict tails from heads and relations.
However, it is often observed that such generative neural models suffer from the diversity and novelty issue: They tend to overproduce high-probability samples similar to those in the training set, while failing to cover a broader range of possible triples absent in a CKG with diverse entities involved. In contrast, discriminatory methods incorporate language models to KG completion tasks such as link prediction and triple classification to evaluate whether a triple is valid, and could be extended to arbitrary text for triples on various KGs \citep{malaviya2019ckgcompletion,davison-etal-2019-commonsense,yao2019kgbert}.
However it would be computationally infeasible to identify new plausible triples on recent large CKGs if we aimlessly explore new nodes and evaluate each possible triple, without leveraging existent nodes and triples as in generative methods. Therefore, all methods above have certain shortcomings.

Particularly, we observe that current methods miss the variation of real world entities, which is a critical factor for the diversity of nodes. For example, a CKG may cover the node \textit{I eat apple} and its relevant edges, but the node \textit{I eat banana} might be missing or the edges are incomplete. As shown in Table~\ref{Tab:CKG_size}, a large portion of the most common real world entities in Probase are never mentioned in the recent CKGs like ASER, not to say edges related to the entities. More, as demonstrated by the low number of average distinct entity per node, directly expanding the scale of CKGs would not be cost-effective to cover diverse entities.

To relieve the issue, we posit the importance of a specific element of human commonsense, \textbf{conceptualization}, which, though found useful for certain natural language understanding tasks \citep{song2011concept,wang2015query,hua2015short}, has not been investigated in depth in this area. As observed by psychologists, ``concepts are the glue that holds our mental world together'' \citep{murphy2004concept}. Human beings are able to make reasonable inferences by utilizing the {\em IsA} relationship between real-world concepts and instances. For example, without knowing what a 
\emph{floppy disk} is, given that it \textbf{is a} \textit{memory device}, people may infer that it may store data and be readable by a computer, etc.
From this viewpoint, instead of directly building triples with countless entities, a CKG can be broadly expanded to handle various queries, as shown in Figure~\ref{Fig:winograd}, by such substitution of instances on text with the corresponding concepts (i.e., \textbf{abstraction}), or vice versa (i.e., \textbf{instantiation}), given an extra CKG of \textit{IsA} relations.

\begin{figure}[t]
\centering
\includegraphics[width=0.95\linewidth]{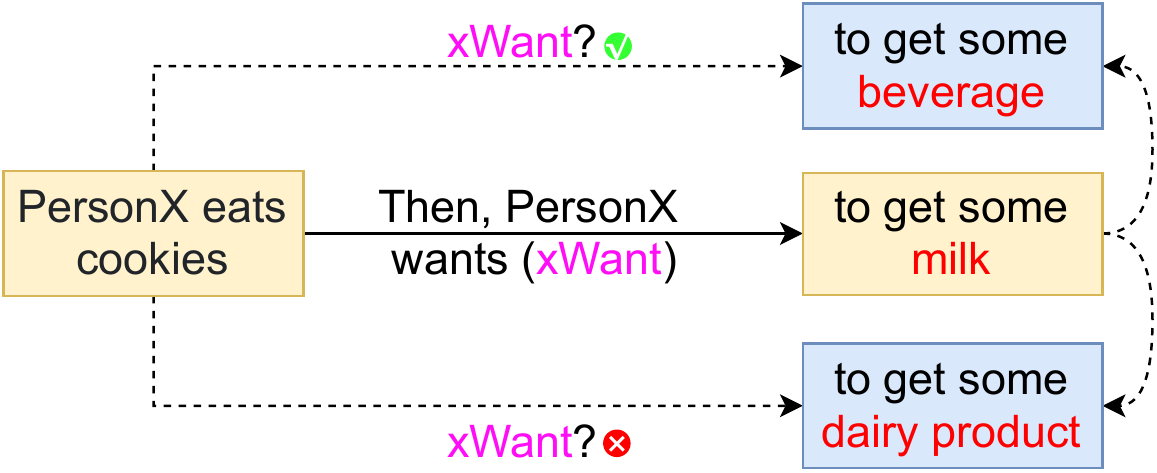}
\caption{A sample from \textsc{Atomic} for discriminating conceptualization. Some conceptualizations, like replacing \textit{milk} with \textit{beverage} in the tail node is valid, while with \textit{dairy} is invalid in the context, as with commonsense one would often want something to drink after eating cookies, while the general concept of \textit{dairy} is not relevant in such a scenario, and dairy products are not all drinkable.}
\label{Fig:concept_sample}
\end{figure}

However, such conceptualization is never strict induction or deduction that is guaranteed to be true. As shown in Figure~\ref{Fig:concept_sample}, it is still a challenging task to determine whether a triple built from conceptualization is reasonable, and requires both the context within the triple and a broader range of commonsense. Such discriminatory problem could be viewed as a particular case of the well-studied task of KG completion.
Therefore, we propose to formulate our problem as a triple classification task, one of the standard tasks for KG completion \citep{socher2013reasoning}. The difference is that, instead of considering arbitrary substitution of the head or tail with existent nodes, we apply conceptualization as described in Section~\ref{ssec:Conceptualization}, and train our model by negative sampling as discussed in Section~\ref{ssec:discriminator}. 
We leverage the rich semantic information with large pretrained language models by fine-tuning them as discriminators. In this way, the models are expected to take triples with arbitrary node as inputs and evaluate whether the triple is reasonable.

To conclude, our contributions are three-fold:

\begin{enumerate}
\item We introduce conceptualization to CKG construction to explore broader range of triples.
\item We formulate the conceptualization-based CKG construction as a triple classification task to be performed on synthetic triples.
\item We propose a method for the task by fine-tuning pretrained language models with negative sampling.
\end{enumerate}
Our code and pipeline 
is available at 
\url{github.com/mutiann/ccc}.

%
%
    %
    %
    %
    %
    %
    %

\section{Methodologies}
\label{sec:methods}

Our methodologies of CKG construction are based on the idea that given a set of ground truth triples as \textit{seeds}, new triples can be built from them by abstraction or instantiation of entities mentioned in the head or tail node, i.e., substituting a mentioned entity with a more general or more specific one, using the particular commonsense of \emph{IsA} relations. Therefore, we need a CKG, \textbf{K}, viewed as seeds, and a conceptualization KG, \textbf{C}, both denoting a set of triples $\langle h,r,t \rangle$, while in \textbf{C}, \emph{r} is always \emph{IsA}.





\subsection{Conceptualization}
\label{ssec:Conceptualization}

Since there are diverse ways to conceptualize an entity from commonsense, \textbf{C} must sufficiently cover various real-world entities connected by \textit{IsA} relations. Therefore, we choose Probase \citep{wu2012probase}, which is a large-scale KG that consists of 17.9M nodes and 87.6M edges, extracted from various corpora, and is proved to be suitable for conceptualization \citep{song2011concept}.

A single entity may have various ways of being abstracted or instantiated, with different typicality. For example, either Linux or BeOS is an \textit{operating system}, and a \textit{pen} is either a writing tool or an enclosure for animals, though for both examples the two choices are not equally common. Since Probase is extracted from real corpora, the frequencies of text showing the triple $\langle h,IsA,t \rangle$in the source corpora can demonstrate how common the relation is. Such
frequencies $f$ are given by Probase along with the triple, forming 4-tuples of $\langle h,IsA,t,f \rangle$. The frequency information in Probase allows us to balance between \textit{IsA} relations of different typicality and filter out noise of rare relations in the graph.

\begin{figure}[H]
\centering
\includegraphics[width=0.95\linewidth]{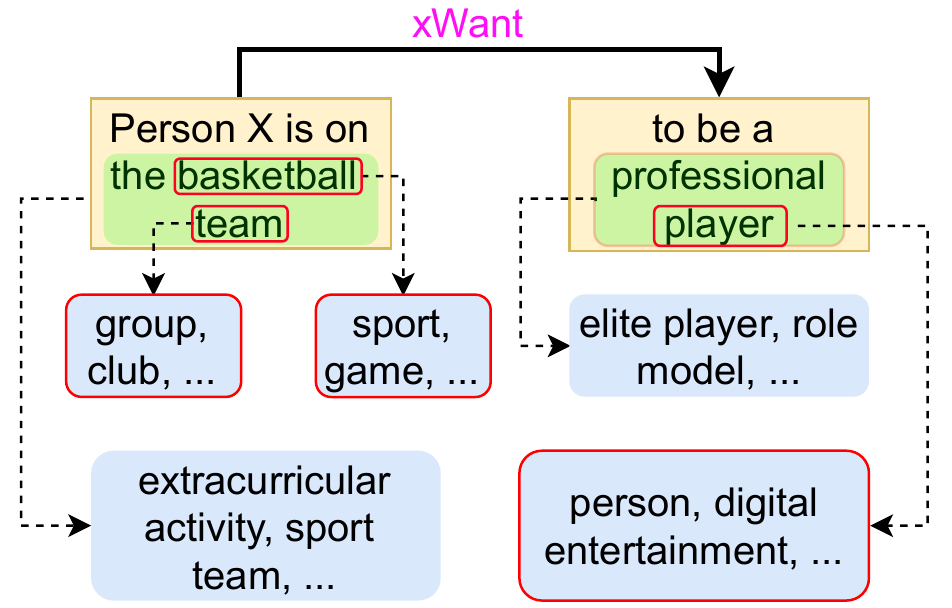}
\caption{A sample of identifying entities. All noun phrases are identified as possible entities on which substitutions are proposed. The phrases include \textit{basketball}, \textit{basketball team}, \textit{team}, \textit{player}, \textit{professional player}, but except \textit{professional} , which is tagged as an adjective here.}
\label{Fig:identify_concepts}
\end{figure}

With \textbf{C} prepared, conceptualization can then be performed on any mentioned real-world entity in the head or tail nodes in each triple, which could be a single noun or noun phrases, serving as subjects, objects or even modifiers, as shown in Figure~\ref{Fig:identify_concepts}. What leads to more complexity is that although nodes in \textbf{C} are all real-world entities in some context, the word or phrase could be used in different manners within a triple of interest. Therefore, for raw text as in \textsc{Atomic}, we choose to perform dependency parsing on each node by spaCy \citep{honnibal2017spacy}. Then we need to filter out all nouns or noun phrases that also present in \textbf{C} as possible candidates.

The method to identify entities is given in Algorithm~\ref{algorithm1}: We iterate through each noun $w$ as the root of the entity, and choose all continuous sequences of words within the range of the subtree corresponding to $w$ in the dependency tree, to ensure that all entities rooted by $w$ (possibly with different modifiers) are collected. We then query the possible abstraction and instantiation of each candidate with Probase, and, for any result, add the substituted text and the corresponding frequency into the list of results. If the text in the CKG are given in the original form (i.e., not lemmatized, unlike ASER), we further use a set of rules to inflect the substitution $s$ returned, and to modify the determiner (if any) in the returned text so as to avoid any false statistical clues of grammatical mistakes introduced by the substitution.

\begin{algorithm}[t]
\SetAlgoLined
\KwIn{\textit{W} $[w_1, w_2, ..., w_n]$: a node represented by a sequence of words  \newline \textit{P} $[p_1, p_2, ..., p_n]$: POS tags of words in \textit{W} \newline \textit{D}: Dependency tree for \textit{W} \newline \textit{C} \{$\langle x, IsA, y, f \rangle$\}: Probase of \textit{IsA} relations}
\KwResult{\textit{S} $[W'_1, W'_2, ...]$: List of substituted word sequences  \newline \textit{F} $[f_1, f_2, ...]$: List of frequencies for each substitution} 
 $\;\;S \gets []$ \\
 $F \gets []$ \\
 \For{$k \in [1, n]$}{
    \If{$p_k \in \{\mathrm{noun,propn}\}$} {
        $\;\;T \gets $subtree of $w_k$ in $D$ \\
        $L \gets \mathrm{min}_{x \in T}$ \{index of $x$ in $W$\} \\
        $R \gets \mathrm{max}_{x \in T}$ \{index of $x$ in $W$\} \\
        \ForEach{$(l, r), L \leq l \leq k \leq r \leq R$}{
            $\;\;E \gets [w_l, ..., w_r]$ \\
                $\;\;A \gets \{(S, f) | (E, IsA, S, f) \in C\}$ \\
                $I \gets \{(S, f) | (S, IsA, E, f) \in C\}$ \\
                \ForEach{$(s, f) \in A \cup I$}{
                    $\;\;S$.add($[w_1, ..., w_{l-1}] + s + [w_{r+1}, ..., w_k]$)  \\
                    $F$.add(f)
                }
            }
    }
 }
\caption{\textsc{IdentifyConceptualization}}
\label{algorithm1}
\end{algorithm}

\subsection{Discriminator}
\label{ssec:discriminator}

We are aimed at building a model capable of evaluating whether a triple is valid, possibly with its head or tail conceptualized. However, as in the well-studied field of KG completion, we fall into the difficult situation that we need to undertake the evaluation using only positive ground truth present in \textbf{K}, except that in our case not only unseen edges but also nodes need to be considered. For such KG completion tasks, commonly it is assumed that it is unknown whether triples not present in \textbf{K} are valid, while the method of negative sampling is applied. In this way, synthetic triples are built from substitution (a.k.a. corruption) of the head or tail of present ones, often using random nodes in the KG. Then the triples are viewed as more likely to be invalid than the original ones, and labelled as members of a negative set $D_{-}$. In combination with triples in \textbf{K} as the positive set $D_{+}$, the model can be trained by such pseudo-labelled data in a self-supervised manner and evaluated by the classification accuracy under triples in \textbf{K} held out from $D_{+}$ and the corresponding negative samples generated in the same way \citep{bordes2013transe,socher2013reasoning,nickel2011rescal}.

Although there could be false-negatives, it is demonstrated that models trained in this way can successfully identify valid triples (though possibly labelled negative) missing from the KG. More, it is discovered that instead of uniform sampling, using negative samples similar to valid ones leads to better performance \citep{cai-wang-2018-kbgan,wang2018igan,zhang2019nscaching}. Therefore, we further propose to sample the substitution of a node from its conceptualized versions, which might be missing in the original KG. This fits into previous KG completion methods if we view the conceptualized new nodes as isolated nodes. We expect that in this way the model could better evaluate conceptualization of triples.

To generate negative samples, two different settings are applied.
\begin{enumerate}
\item Node Substitution (\textbf{NS}): The common corruption method, as in \citet{bordes2013transe}, that substitutes the head or tail (each with 0.5 probability) with a random node from the KG. For a CKG like \textsc{Atomic} in which head nodes and tails nodes, as well as tail nodes from triples with different relations, can often be easily distinguished from each other\footnote{For instance, the head node in \textsc{Atomic} always starts with \textit{Person X}, and tail nodes for relation type \textit{xAttr} (attribute of Person X) are often adjectives.}, we follow \citet{socher2013reasoning} to pick random heads only from other heads, and random tails only from other tails appearing in triples with the same relation.
\item Entity Conceptualization (\textbf{EC}): To enable the model to identify false triples with inappropriate conceptualization, we randomly choose the head or tail (each with 0.5 probability) and corrupt the node as in Section~\ref{ssec:Conceptualization} by substituting an entity in the node with its abstraction or instantiation. This method ensures that the substituted nodes are often plausible. Then we use the triples with the head or tail substituted as negative samples. Particularly, we make use of the frequencies returned by Algorithm~\ref{algorithm1} as weights (or unnormalized probabilities), based on which we sample from possible conceptualized nodes, as shown in Algorithm~\ref{ref:algorithm2}. In this way, we strike a balance between the diversity and the typicality of \emph{IsA} relations used.
\end{enumerate}



\begin{algorithm}[H]
\SetAlgoLined
\KwIn{\textit{N}: A node, represented by a sequence of words\newline \textit{P}: POS tags of words in N \newline \textit{D}: Dependency tree for \textit{N} \newline \textit{C}: Probase}
\KwResult{$N'$: corrupted node}
$\;\;S$, $F$ $\gets$ \textsc{IdentifyConceptualization}($N$, $P$, $D$, $C$)\\
$W$ $\gets$ $F$ / $\sum F$ \\
$k \sim Categorical(W)$ \\
$N'$ $\gets$ $S_k$
\caption{\textsc{BuildSampleEC}}\label{ref:algorithm2}
\end{algorithm}
Building negative samples with both settings, we expect that the model will be capable of discriminating whether a triple is valid or not, when the triple is possibly corrupted in either way. 
To reduce noise in training and evaluation, negative samples are filtered to ensure that they are different from any positive samples, which matches the \textit{filtered} setting from \citet{bordes2013transe}.

\begin{figure}[H]
\centering
\includegraphics[width=0.95\linewidth]{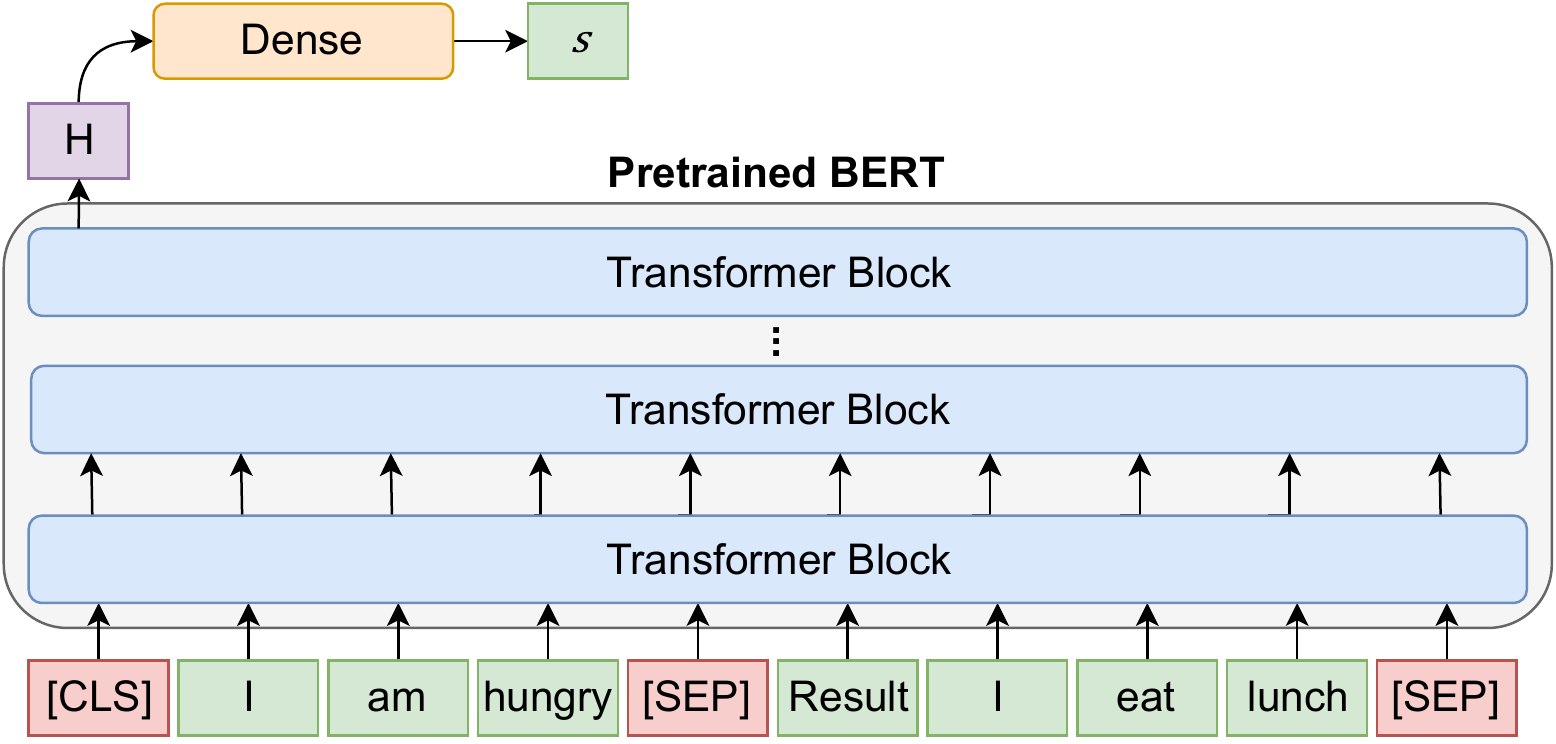}
\caption{Architecture of the BERT-based discriminator model. Raw text are fed into the model to predict the binary label \textit{y}. All except the last fully-connected layer are pretrained but not frozen.}
\label{Fig:model_arch}
\end{figure}

To make the best use of semantic information from the textual description of nodes, we apply the widely used transformer-based pretrained language models like BERT as the discriminator. Particularly, the structure of our task matches the next sentence prediction (NSP) in \citet{devlin2018bert}. As a result, we follow a similar setting that takes pairs of sentences separated by a special [SEP] token and marked by different token type IDs as inputs, with \emph{h} the first sentence and the concatenation of \emph{r} and \emph{t} the second sentence. Binary classification is then performed by a fully-connected layer taking the final representation corresponding to the [CLS] token, as shown in Figure~\ref{Fig:model_arch}. All parameters in the model, except those in the final fully-connected layer for binary classification, can be initialized from the pretrained  model. Then the model is fine-tuned using the positive and negative samples mentioned above, with a 1:1 frequency during training, using binary cross-entropy loss below, based on the output $s$, which is a scalar after a logistic sigmoid activation, indicating the confidence on the input being valid:

\begin{equation}
L=-\sum_{(x,y) \in D_{+} \cup D_{-}} (y\log s+(1-y)\log(1-s)).
\end{equation}

\section{Experiments}
\subsection{Datasets}
Two different datasets, \textsc{Atomic} and ASER, which are typical CKGs using open-form text as nodes, are used in our experiments. Earlier CKGs such as ConceptNet \citep{speer2017conceptnet} are not discussed, as in ConceptNet, unlike the more recent CKGs, only simple text, mostly noun phrases, is used as nodes, and previous work can already reach close-to-human results \citep{saito-etal-2018-commonsense}.

\subsubsection{\textsc{Atomic}}
\textsc{Atomic} is a CKG of 877K triples on cause-and-effect relations between everyday activities \citep{sap2019Atomic}. Within \textsc{Atomic}, head nodes, or base events, are extracted from text corpora, with the form of some person's action or state (e.g., \textit{Person X is on the basketball team}). The dataset further categorizes the cause-and-effect relations into nine types, covering the intention, prerequisites and impacts with regard to the agent \textit{Person X} and other people. The tail entities are then written in open-form text by crowd-sourced workers under these categories. In this way, a broad range of eventualities and their relations are covered by the dataset. 
We follow the original data split of \textsc{Atomic} in our experiments.

\subsubsection{ASER}
ASER \citep{zhang2019aser} is a large-scale CKG of 194M nodes representing verb-centric eventualities matching certain patterns (e.g., \textit{s-v-o} for \textit{I love dogs} and \textit{s-v-v-o} for \textit{I want to eat an apple}) extracted from various corpora. Relations of 15 types between nodes are then extracted as well, identified by matching the text with language patterns such as ``$E_1$, because $E_2$'' for \textbf{Reason} and ``$E_1$, $E_2$ instead'' for \textbf{ChosenAlternative}. A total 194.0M nodes and 64.4M triples are extracted in ASER.
In our experiments, we use the core release of ASER with 27.6M nodes and 10.4M edges. Triples of the \textbf{CoOccurance} type and isolated nodes are further removed to create a smaller and cleaner KG with 1.4M nodes and 1.1M edges. Triples are then randomly split into \textit{train}, \textit{dev}, and \textit{test} set at 8:1:1.

\subsection{Settings}
To build our CKG Construction by Conceptualization (\textbf{CCC}) discriminator, we follow the scheme for fine-tuning BERT on downstream tasks \citep{devlin2018bert}, and use the pretrained 12-layer BERT-base model on GTX1080 GPUs with 8GB memory. To evaluate the impact of the two different ways of producing negative samples given in Section~\ref{ssec:discriminator}, and to trade off between model capabilities of discriminating triples under general cases and specifically identifying inappropriate conceptualization, we perform experiments with different percentages of negative samples built by conceptualization, i.e., the EC setting. Specifically, models with 50\%, 75\%, and 87.5\% negative samples created by EC (and the rest by NS) are trained and reported. For evaluation, negative samples are generated on the \textit{dev} and \textit{test} samples as well by both methods, forming the EC and NS \textit{dev} and \textit{test} sets with 1:1 positive and negative samples. Under EC, those triples with nodes containing no entities to be conceptualized (e.g., \textit{I am fine}, for which Algorithm~\ref{algorithm1} returns empty results) are ignored. Nevertheless, 79.65\% and 83.56\% of triples in the \textit{dev} and \textit{test} sets are collected in the EC set for \textsc{Atomic} and ASER respectively, showing that a majority of triples can be conceptualized. \textit{Test} results of the models at best EC \textit{dev} accuracy are reported.

\subsection{Baselines}
We train COMeT\footnote{Available at \url{github.com/atcbosselut/comet-commonsense}} and KG-BERT\footnote{Available at \url{github.com/yao8839836/kg-bert}} on the two datasets as our baselines. Particularly, our model will degenerate into KG-BERT with 0\% EC samples, as the NS setting is what KG-BERT trained under. Since COMeT itself is not a discriminative model, we use the perplexity per token\footnote{Unlike the case in \citet{malaviya2019ckgcompletion}, conceptualization would not significantly change the length of a triple, so we only use the \textsc{Normalized} setting.} as its score given to each triple, and use the \textit{dev} set to find a classification threshold with best accuracy.

\begin{table*}[t!]
\centering
\begin{tabular}{lcccc}
\toprule
	& \multicolumn{2}{c}{\textbf{ASER}} & \multicolumn{2}{c}{\textbf{\textsc{Atomic}}} \\
	 & \textsc{EC} & \textsc{NS} & \textsc{EC} & \textsc{NS} \\ \midrule
	COMeT & 0.6388 & 0.5869 & 0.6927 & 0.5730 \\
	KG-BERT & 0.7091 & \textbf{0.8018} & 0.7669 & 0.6575 \\ \hline
	CCC-50 & 0.8716 & 0.7775 & 0.9016 & \textbf{0.7840} \\
	CCC-75 & 0.8995 & 0.7250 & 0.9221 & 0.7446 \\
	CCC-87.5 & \textbf{0.9156} & 0.6635 & \textbf{0.9355} & 0.6980 \\ \hline
	CCC-75-scratch & 0.8284 & 0.5587 & 0.8579 & 0.5003 \\
	CCC-75-RoBERTa & 0.8999 & 0.6938 & 0.9305 & 0.7350 \\ \bottomrule
\end{tabular}

\caption{\label{Tab:accuracy} Results of accuracy on EC and NS \textit{test} set of baselines and our models on different datasets. \textbf{CCC} denotes our model, with the number attached representing percentage of EC training.}
\end{table*}

\subsection{Results}
\subsubsection{Triple Classification}
Test accuracies of triple classification on both methods are given in Table~\ref{Tab:accuracy}. As shown by the results, COMeT lacks discriminative power, which is consistent with the results in \citet{malaviya2019ckgcompletion}. KG-BERT, which has been successfully applied on traditional KGs, produce satisfactory results on CKG as well, while our methods perform better than both baseline methods by a large margin on the EC tests. Hence it is demonstrated that introducing conceptualization during training is effective to create a model capable of identifying false conceptualization. Particularly, the percentage of EC samples in training is critical for a trade-off between EC and NS tasks: Increased EC percentage will lead to better EC results, but the NS results will drop. The \textsc{Atomic} CCC models reach better results on NS than KG-BERT, which is possibly due to the fact that \textsc{Atomic} nodes are mostly about everyday activities, in contrast to ASER which covers a broader range of topics. Therefore by EC training a more diverse set of nodes could be seen by the model in training, and could be helpful for the model to generalize in the NS test.

\subsubsection{Ablation Studies}
We perform ablation studies to examine the importance of pretraining and model selection. With the model trained from scratch on our task without using pretrained parameters, the performance significantly drops, as shown in the CCC-75-scratch results in Table~\ref{Tab:accuracy}. We also attempted to use RoBERTa, an alternative pretrained language model that makes improvements on BERT training and has demonstrated better performance on downstream tasks  \citep{liu2019roberta}. However, the results using the pretrained RoBERTa-base model (CCC-75-RoBERTa) are generally on par with our model using BERT. This could be possibly explained by that BERT is sufficient in our current settings, that RoBERTa uses a larger batch size while on our GPU the batch size is more limited, and that the NSP pretraining task is used in BERT but is absent in RoBERTa, as NSP exactly matches the input scheme of our task.

\begin{table*}[t]
\centering
\begin{tabular}{lcccc}
\toprule
	&  \multicolumn{2}{c}{\textbf{ASER}} &  \multicolumn{2}{c}{\textsc{\textbf{\textsc{Atomic}}}} \  \\ 
	& COMeT & CCC-75 & COMeT & CCC-75 \  \\ \midrule
	N/Seed & 10 & 8.28 & 10 & 5.02 \\ \hline
	Dist-N & 24.68\% & 96.57\% & 6.49\% & 51.26\% \\ 
	Dist-1 & 1.62\% & 6.30\% & 0.63\% & 8.34\% \\ 
	Dist-2 & 10.56\% & 15.45\% & 2.87\% & 49.76\% \\ \hline
	N/T N & 88.48\% & 98.60\% & 10.30\% & 94.65\% \\ 
	N/U N & 93.38\% & 99.18\% & 69.17\% & 96.66\% \\ \hline
	Dist-N-Norm & 10.74\% & 84.16\% & 4.35\% & 46.36\% \\ 
	N/T N-Norm & 9.37\% & 86.01\% & 5.12\% & 87.95\% \\ 
	N/U N-Norm & 65.72\% & 95.02\% & 58.71\% & 93.01\% \\ \bottomrule
\end{tabular}

\caption{\label{Tab:diversity} Results for diversity and novelty on generations, larger for better results. All rows except N/seed are given in percentage.}
\end{table*}

\begin{table*}[t]
\centering
\begin{tabular}{lll}
\toprule
 & \multicolumn{1}{c}{\textit{head}}  & \multicolumn{1}{c}{\textit{tail}} \\
\midrule
Seed & another promises him a scholarship & his parents own a successful business \\ 
\hline
\multirow{5}{4em}{COMeT} & \multicolumn{1}{c}{\multirow{5}{*}{--}} & he never gets it\\
& & he does not get it\\
& & he never gets one\\
& & he could not pay it\\
& & he does not receive it\\ \hline

\multirow{5}{4em}{CCC-75} & another promises him a grant & \multicolumn{1}{c}{\multirow{2}{*}{--}}\\
& another promises him an award &\\
& \multicolumn{1}{c}{\multirow{3}{*}{--}} & his parents own a successful shop\\
& & his parents own a successful bank\\
& & his parents own a successful hotel\\
\bottomrule
\end{tabular}

\caption{\label{Tab:generation} Samples for ASER generations given the seed. In this sample the head and tail are connected by the relation of \textbf{Concession}, i.e., \textit{although}.}
\end{table*}
\subsubsection{Generations}

We generate triples using the \textit{test} set from ASER and \textsc{Atomic} as seeds by both COMeT with 10-beam search and the CCC-75 model by conceptualization, and apply various metrics on the results, as shown in Table~\ref{Tab:diversity}. Both methods may produce a large number of triples, as given by the number of generations per seed, \textbf{N/Seed}. For diversity, we report \textbf{Dist-1} (number of distinct words per node), \textbf{Dist-2} (number of distinct bigrams per node), and \textbf{Dist-N} (number of distinct nodes per node). Due to the different number of generated triplets, the results are all normalized by number of nodes.\footnote{Results from  \citet{bosselut-etal-2019-comet} of test perplexity are reproduced in our experiments. Differences on diversity metrics are due to the fact that we use 10-beam search instead for fair comparison.}

Novelty is measured by \textbf{N/T N}, the proportion of nodes among all produced nodes that are novel, i.e. not present in the training set, and \textbf{N/U N}, the proportion of novel distinct nodes, among all distinct nodes. Moreover, since generative methods may produce nodes of essentially the same meaning with slight changes in forms, we also have the produced nodes normalized, by removing structural words like determiners, auxiliary verbs, pronouns, etc. We then report the results for the metrics above but applied to generations after such normalization, denoted as \textbf{Dist-N-Norm}, \textbf{N/T N-Norm}, and \textbf{N/U N-Norm} respectively. Furthermore, samples of generations by both models are shown in Table~\ref{Tab:generation}.

It is clearly demonstrated in the diversity results that a majority of generations by COMeT are similar to each other given a certain head node and relation, as the number of distinct nodes, words and bigrams are all relatively low. The novelty results further show that generated nodes are often similar to the seen ones in the training set as well. It can be particularly observed that, though the original diversity and novelty metrics appear to be acceptable, which is consistent with \citet{bosselut-etal-2019-comet}, results drop sharply when the generations are normalized. This indicates that COMeT may produce slightly different nodes paraphrasing each other, as shown in Table~\ref{Tab:generation} where four of five generated tails are similar to each other (saying \textit{he does not get it}), while this is not the case for CCC, as generated nodes are mostly discussing different entities, and the results will often be diverse and novel.

\section{Related Work}

Automatic construction of structured KGs is a well-studied task, and a number of learning-based methods have been proposed, including KG embedding methods based on translational distances \citep{bordes2013transe,lin2015transr,ji-etal-2015-knowledge,wang2014transh,shang2019convtranse} and semantic matching \citep{nickel2011rescal,socher2013reasoning,yang2014distmult,trouillon2016complex}, typically trained by negative sampling techniques and applied on tasks like link prediction and triple classification. Furthermore, graph neural networks can be used to better capture structural information \citep{schlichtkrull2018rgcn}, GANs are applied to improve negative sampling \citep{cai-wang-2018-kbgan,wang2018igan,zhang2019nscaching} by mining more difficult examples, and textual information from the node can be leveraged\citep{wang2016teke,xie2016representation,xiao2017ssp,an-etal-2018-accurate}. 

Textual information is more critical on CKGs with nodes carrying complicated eventualities, often in open-form text. Therefore, \citet{li-etal-2016-commonsense} proposed to score ConceptNet triples by neural sequence models taking text inputs so as to discover new triples, while \citet{saito-etal-2018-commonsense} and \citet{sap2019Atomic} further proposed to generate tail nodes by a sequence-to-sequence LSTM model with head and relation as inputs. Recently powerful large pretrained models like BERT and GPT-2 have been proposed \citep{devlin2018bert,radford2019gpt2}, from which, it has been observed by \citet{trinh2018simple} and \newcite{radford2019gpt2} that rich knowledge including commonsense can be extracted. Therefore, different ways for KG construction have been introduced on such models as downstream tasks: In KG-BERT, BERT was fine-tuned for KG completion tasks like link prediction and triple classification \citep{yao2019kgbert}; COMeT used GPT-based models to generate tails \citep{bosselut-etal-2019-comet}; LAMA directly predicted masked words in triples on various KGs by BERT \citep{petroni-etal-2019-language}; \citet{davison-etal-2019-commonsense} considered both generation of new tails and scoring given triples; \citet{malaviya2019ckgcompletion} utilized both structural and semantic information for CKG construction on link prediction tasks.

\section{Conclusion}
We introduce conceptualization to commonsense knowledge graph construction and propose a novel method for the task by generating new triples with conceptualization and examine them by a discriminator transferred from pretrained language models. Future studies will be focused on strategies of conceptualization and its role in natural languages and commonsense by deep learning approaches.

\bibliography{anthology,emnlp2020}
\bibliographystyle{acl_natbib}

\end{document}